\documentclass[11pt,a4paper]{article}
\usepackage[hyperref]{eacl2021}
\usepackage{times}
\usepackage{latexsym}
\usepackage{multirow}
\usepackage{color}
\usepackage{tcolorbox}
\usepackage{CJK}
\usepackage{adjustbox}

\aclfinalcopy 

\title{IIITT@LT-EDI-EACL2021-Hope Speech Detection: There is always Hope in Transformers}

\author{Karthik Puranik\(^1\), Adeep Hande\(^1\), Ruba Priyadharshini\(^2\), \\\textbf{Sajeetha Thavareesan\(^3\)}, \textbf{Bharathi Raja Chakravarthi\(^4\)} \\  \(^1\)Indian Institute of Information Technology Tiruchirappalli \\ \(^2\)ULTRA Arts and Science College, India,\hfill \(^3\)Eastern University, Sri Lanka\\  \(^4\)National University of Ireland Galway \\  {\tt karthikp18c@iiitt.ac.in} } 

\date{}

\usepackage{graphicx}
\begin{document}
\maketitle
\begin{abstract}
In a world filled with serious challenges like climate change, religious and political conflicts, global pandemics, terrorism, and racial discrimination, an internet full of hate speech, abusive and offensive content is the last thing we desire for. In this paper, we work to identify and promote positive and supportive content on these platforms. We work with several transformer-based models to classify social media comments as hope speech or not-hope speech in  English, Malayalam and Tamil languages. This paper portrays our work for the Shared Task on Hope Speech Detection for Equality, Diversity, and Inclusion at LT-EDI 2021- EACL 2021. The codes for our best submission can be viewed\footnote{\url{https://github.com/karthikpuranik11/Hope-Speech-Detection-}}.
\end{abstract}

\section{Introduction}

Social Media has inherently changed the way people interact and carry on with their everyday lives as people using the internet \cite{9074205,9074379}. Due to the vast amount of data being available on social media applications such as YouTube, Facebook, an Twitter it has resulted in people stating their opinions in the form of comments that could imply hate or negative sentiment towards an individual or a community \cite{10.1145/3441501.3441515,10.1145/3441501.3441517}. This results in people feeling hostile about certain posts and thus feeling very hurt \cite{bhardwaj2020hostility}. 

Being a free platform, social media runs on user-generated content. With people from multifarious backgrounds present, it creates a rich social structure \cite{article3} and has become an exceptional source of information. It has laid it's roots so deeply into the lives of people that they count on it for their every need. Regardless, this tends to mislead people in search of credible information. Certain individuals or ethnic groups also fall prey to people utilizing these platforms to foster destructive or harmful behaviour which is a common scenario in cyberbullying \cite{doi:10.1080/02673843.2019.1669059}.

\begin{table*}[!h]
 
\begin{center}
\scalebox{0.87}{
\begin{tabular}{|l|r|r|}
\hline

\textbf{Text} & \textbf{Language} & \textbf{Label}\\
\hline
God gave us a choice my choice is to love,  I would die for that kid  & English & Hope \\
\hline
The Democrats are.Backed powerful rich people like Soros &  English & Not hope \\
\hline
ESTE PSICÃ“PATA MASÃ“N LUCIFERIANO ES HOMBRE TRANS & English & Not English\\
\hline
Neega podara vedio nalla iruku ana subtitle vainthuchu ahh yella language papaga  & Tamil & Hope \\
\hline
Avan matum enkita maatunan... Avana kolla paniduven & Tamil & Not hope \\
\hline
I can't uninstall mY Pubg & Tamil & Not Tamil\\
\hline
ooororutharum avarude ishtam pole jeevikatte . k. & Malayalam & Hope \\ 
\hline
Etraem aduthu nilkallae Arunae & Malayalam & Not hope \\
\hline
Phoenix contact me give you’re mail I’d I hope I can support you sure! & Malayalam & Not Malayalam\\
\hline
\end{tabular}
}\caption{Examples of hope speech or not hope speech}\label{tab3}
\end{center}
\end{table*}
The earliest inscription in India dated from 580 BCE was the Tamil inscription in pottery and then, the Asoka inscription in Prakrit, Greek and Aramaic dating from 260 BCE. Thunchaththu Ramanujan Ezhuthachan split Malayalam from Tamil after the 15th century CE by using Pallava Grantha script to write religious texts. Pallava Grantha was used in South India to write Sanskrit and foreign words in Tamil literature. Modern Tamil and Malayalam have their own script.  However, people use the Latin script to write on social media \cite{chakravarthi-etal-2018-improving,chakravarthi-etal-2019-wordnet,chakravarthi2020leveraging}. 

The automatic detection of hateful, offensive, and unwanted language related to events and subjects on gender, religion, race or ethnicity in social media posts is very much necessary \cite{rizwan-etal-2020-hate,nikhiloffen,nikhilhope}. Such harmful content could spread, stimulate, and vindicate hatred, outrage, and prejudice against the targeted users. Removing such comments was never an option as it suppresses the freedom of speech of the user and it is highly unlikely to stop the person from posting more. In fact, he/she/they would be prompted to post more of such comments\footnote{https://www.qs.com/negative-comments-on-social-media/} \cite{adeepoffensive,adeepmeme}. This brings us to our goal to spread positivism and hope and identify such posts 
to strengthen an open-minded, tolerant, and unprejudiced society.

\section{Related Works}
The need for the segregation of toxic comments from social media platforms has been identified back in the day. \citet{founta2018unified} has tried to study the textual properties and behaviour of abusive postings on Twitter using a Unified Deep Learning Architecture. Hate speech can be classified into various categories like hatred against an individual or group belonging to a race, religion, skin colour, ethnicity, gender, disability, or nation\footnote{\url{http://www.ala.org/advocacy/intfreedom/hate} (Accessed January 16, 2021)} and there have been studies to observe it's evolution in social media over the past thirty years \cite{RePEc:spr:scient:v:126:y:2021:i:1:d:10.1007_s11192-020-03737-6}. Deep Learning methods were used to classify hate speech into racist, sexist or neither in \citet{Badjatiya_2017}.

Hope is support, reassurance or any kind of positive reinforcement at the time of crisis \cite{chakravarthi-2020-hopeedi}. \citet{palakodety2020hope} identifies the need for the automatic detection of content that can eliminate hostility and bring about a sense of hope during times of wrangling and brink of a war between nations. There have also been works to identify hate speech in multilingual \cite{aluru2020deep} and code-mixed data in Tamil, Malayalam, and Kannada language \cite{chakravarthi-etal-2020-corpus,chakravarthi-etal-2020-sentiment,hande-etal-2020-kancmd}. However, there have been very fewer works in Hope speech detection for Indian languages.

\begin{figure*}[!h]
\centering
\includegraphics[width=12cm,height =8cm]{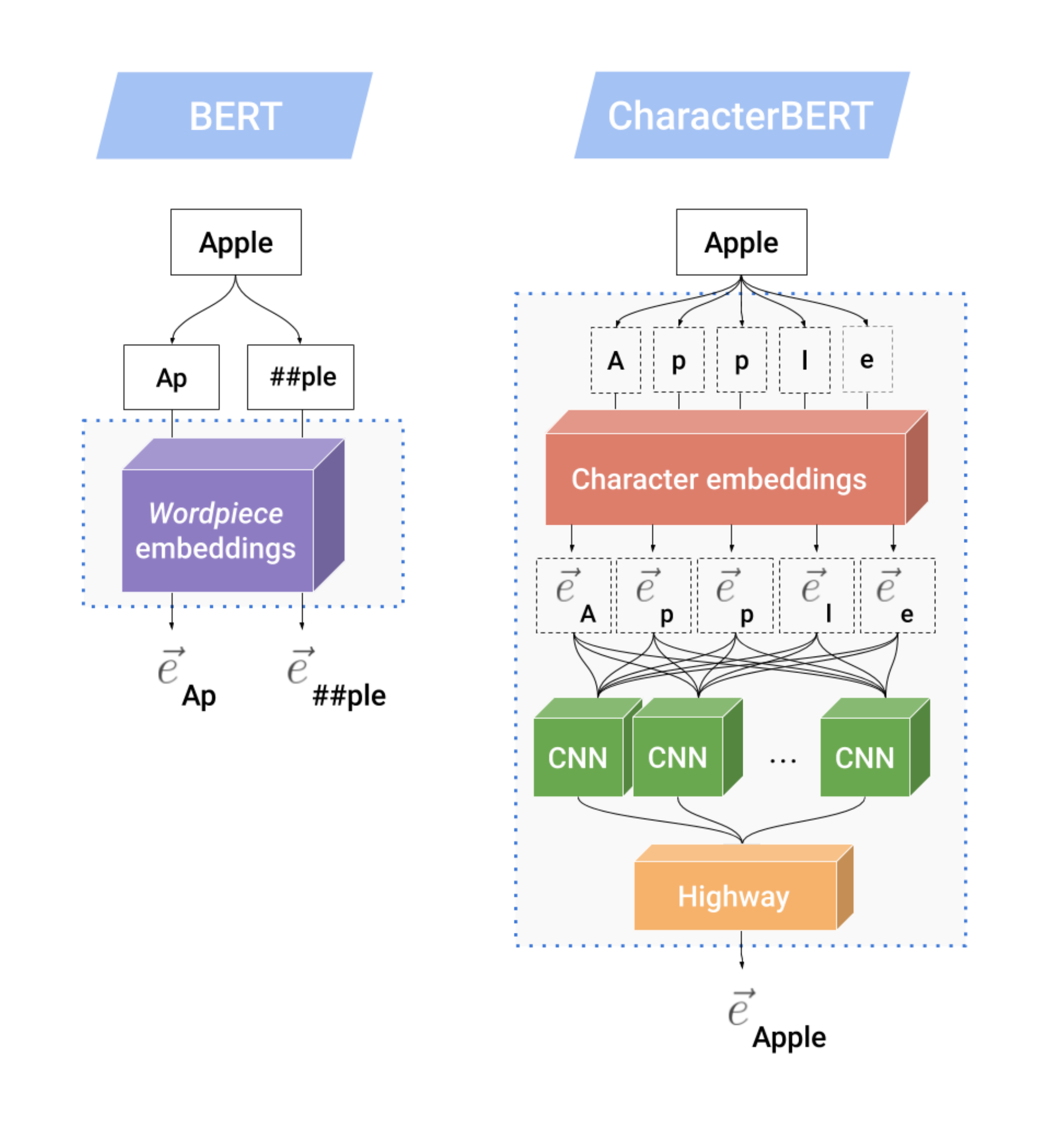}
\caption{Context-independent representations in BERT and CharacterBERT (Source: \citet{el-boukkouri-etal-2020-characterbert})} \label{fig1}
\end{figure*} 

\section{Dataset}
The dataset is provided by \cite{chakravarthi-2020-hopeedi} \cite{dravidianhopespeech-eacl} and contains 59,354 comments from the famous online video sharing platform YouTube out of which 28,451 are in English, 20,198 in Tamil, and 10,705 comments are in Malayalam (Table \ref{tab1}) which can be classified as Hope speech, not hope speech and other languages. This dataset is split into train (80\%), development (10\%) and test (10\%)  dataset (Table \ref{tab2}). 

Subjects like hope speech might raise confusions and disagreements between annotators belonging to different groups. The dataset was annotated by a minimum of three annotators and the inter-annotator agreement was determined using Krippendorff’s alpha \cite{article4}. Refer table \ref{tab3} for examples of hope speech, not hope speech and other languages for English, Tamil and Malayalam datasets respectively.

\begin{table}[!h]
\begin{center}
    
\renewcommand{\tabcolsep}{3mm}
\scalebox{0.93}{
\begin{tabular}{|l|r|r|r|}
\hline
Class &  English & Tamil & Malayalam\\
\hline

Hope &  2,484 & 7,899 & 2,052\\
 
Not Hope & 25,940 & 9,816 & 7,765\\
Other lang & 27 & 2,483 & 888\\
\hline
Total & 28,451 & 20,198 & 10,705\\
\hline
\end{tabular}
}
\end{center}
\caption{ Classwise Data Distribution}\label{tab1}
\end{table}

\begin{table}[!h]
\begin{center}
\renewcommand{\tabcolsep}{3mm}
 \scalebox{0.9}{
\begin{tabular}{|l|r|r|r|}
\hline

Split &  English & Tamil & Malayalam\\
\hline
Training &  22,762 & 16,160 & 8564\\
Development & 2,843 & 2,018 & 1070\\
Test & 2,846 & 2,020 & 1071\\
\hline
Total & 28,451 & 20,198 & 10,705\\
\hline
\end{tabular}
}
\end{center}
\caption{ Train-Development-Test Data Distribution}\label{tab2}
\end{table}

\section{Experiment Setup}
In this section, we give a detailed explanation of the experimental conditions upon which the models are developed.
\subsection{Architecture}
\subsubsection{Dense}
The dense layers used in CNN (convolutional neural networks) connects all layers in the next layer with each other in a feed-forward fashion \cite{huang2018densely}. Though they have the same formulae as the linear layers i.e. wx+b, the output is passed through an activation function which is a non-linear function. We implemented our models with 2 dense layers, rectified linear units (ReLU) \cite{agarap2019deep} as the activation function and dropout of 0.4. 
\subsubsection{Bidirectional LSTM}
Bidirectional LSTM or biLSTM is a sequence processing model \cite{650093}. It uses both the future and past input features at a time as it contains two LSTM's, one taking input in the forward direction and another in the backward direction~\cite{Schuster1997BidirectionalRN}. The backward and forward pass through the unfolded network just like any regular network. However, BiLSTM requires us to unfold the hidden states for every time step. It produces a drastic increase in the size of information being fed thus, improving the context available~\cite{huang2015bidirectional}. Refer Table \ref{params} for the parameters used in the BiLSTM model.

\begin{table}[htbp]
 
\begin{center}
    
\renewcommand{\tabcolsep}{3mm}
\scalebox{1}{
\begin{tabular}{|l r|} 
 \hline
 \textbf{Parameter}& \textbf{Value}\\ 
 \hline
 
Number of LSTM units & 256 \\ 
Dropout & 0.4 \\
Activation Function & ReLU \\
Max Len & 128 \\
Batch Size & 32 \\
Optimizer & AdamW\\
Learning Rate & 2e-5\\
Loss Function & cross-entropy\\
Number of epochs & 5\\
 
\hline
\end{tabular}
}
\end{center}
\caption{Parameters for the BiLSTM model }
\label{params}
\end{table}

\subsection{Embeddings}
\subsubsection{BERT}
Bidirectional Encoder Representations from Transformers (BERT) \cite{devlin-etal-2019-bert}. The multilingual base model is pretrained on the top 104 languages of the world on Wikipedia (2.5B words) with 110 thousand shared wordpiece vocabulary. The input is encoded into vectors with BERT's innovation of bidirectionally training the language model which catches a deeper context and flow of the language. Furthermore, novel tasks like Next Sentence Prediction (NSP) and Masked Language Modelling (MLM) are used to train the model. 

The pretrained BERT Multilingual model \textbf{\emph{bert-base-multilingual-uncased}} \cite{pires-etal-2019-multilingual} from Huggingface\footnote{\url{https://huggingface.co/transformers/pretrained_models.html}} \cite{wolf2020huggingfaces} is executed in PyTorch \cite{NEURIPS2019_bdbca288}. It consists of 12-layers, 768 hidden, 12 attention heads and 110M parameters which are fine-tuned by concatenating with bidirectional LSTM layers. The BiLSTM layers take the embeddings from the transformer encoder as the input which increases the information being fed, which in turn betters the context and accuracy. Adam algorithm with weight decay fix is used as an optimizer. We train our models with the default learning rate of $2e-5$. We use the cross-entropy loss as it is a multilabel classification task.

\begin{table*}[htbp]
 
\begin{center}
\begin{tabular}{|c|r|r|r|}
\hline
\textbf{Architecture} & \textbf{{Embeddings}} & \textbf{F1-Score validation} & \textbf{F1-Score test}\\
\hline
\hline
BiLSTM & bert-base-uncased & 0.9112 & 0.9241 \\
\hline
\multirow{5}{4em}{Dense} & bert-base-uncased & 0.9164 &0.9240 \\
\cline{2-4}
& albert-base & 0.9143 &0.9210 \\
\cline{2-4}
& distilbert-base-uncased & 0.9238 &0.9283 \\
\cline{2-4}
& roberta-base & 0.9141 &0.9235 \\
\hline
& character-bert & \textbf{0.9264} & 0.9220\\
\hline
& ULMFiT & 0.9252 & 0.9356 \\
\hline
\end{tabular}
\caption{Weighted F1-scores of hope speech detection classifier models on English dataset}
\label{tab6}
\end{center}
\end{table*}

\begin{table*}[!h]
 
\begin{center}
\begin{tabular}{|c|r|r|r|}
\hline
\textbf{Architecture} & \textbf{{Embeddings}} & \textbf{F1-Score validation} & \textbf{F1-Score test}\\
\hline
\hline
\multirow{4}{4em}{BiLSTM} & mbert-uncased & \textbf{0.8436} & 0.8545 \\
\cline{2-4}
& mbert-cased & 0.8280 & 0.8482 \\
\cline{2-4}
& xlm-roberta-base & 0.8271 & 0.8233 \\
\cline{2-4}
& MuRIL & 0.8089 & 0.8212\\
\hline
\multirow{5}{4em}{Dense} & mbert-uncased & 0.8373 & 0.8433 \\
\cline{2-4}
& indic-bert & 0.7719 & 0.8264 \\
\cline{2-4}
& xlm-roberta-base & 0.7757 & 0.7001 \\
\cline{2-4}
& distilmbert-cased & 0.8312 & 0.8395 \\
\cline{2-4}
& MuRIL & 0.8023 & 0.8187\\
\hline
\end{tabular}
\caption{Weighted F1-scores of hope speech detection classifier model on Malayalam dataset}
\label{tab7}
\end{center}
\end{table*}

\subsubsection{ALBERT}

It has a similar architecture as that of BERT but due to memory limitations and longer training periods, ALBERT or A Lite BERT introduces two parameter reduction techniques \cite{chiang-etal-2020-pretrained}. ALBERT distinguishes itself from BERT with features like factorization of the embedding matrix, cross-layer parameter sharing and inter-sentence coherence prediction. We implemented \textbf{\emph{albert-base-v2}} pretrained model with 12 repeating layers, 768 hidden, 12 attention heads, and 12M parameters for the English dataset.

\subsubsection{DistilBERT}
DistilBERT is a distilled version of BERT to make it smaller, cheaper, faster, and lighter \cite{unknown5}. With up to 40\% less number of parameters than \textbf{\emph{bert-base-uncased}}, it promises to run 60\% faster while preserving 97\% of it's performance. We employ \textbf{\emph{distilbert-base-uncased}} for the English dataset and \textbf{\emph{distilbert-base-multilingual-cased}} for the Tamil and Malayalam datasets. Both models have 6-layers, 768-hidden, 12-heads and while the former has 66M parameters, the latter has 134M parameters.

\subsubsection{RoBERTa}

A Robustly optimized BERT Pretraining Approach (RoBERTa) is a modification of BERT \cite{liu2020roberta}. RoBERTa is trained for longer, with larger batches on 1000\% more data than BERT. The Next Sentence Prediction (NSP) task employed in BERT's pre-training is removed and dynamic masking during training is introduced. It's additionally trained on a 76 GB large new dataset (CC-NEWS). \textbf{\emph{roberta-base}} follows the BERT architecture but has 125M parameters and is used for the English dataset.

\begin{table*}[!h]
 
\begin{center}
\begin{tabular}{|c|r|r|r|}
\hline
\textbf{Architecture} & \textbf{{Embeddings}} & \textbf{F1-Score Validation} & \textbf{F1-Score test}\\
\hline
\hline
\multirow{4}{4em}{BiLSTM} & mbert-uncased & 0.6124 & 0.5601 \\
\cline{2-4}
& mbert-cased & \textbf{0.6183} & 0.5297 \\
\cline{2-4}
& xlm-roberta-base & 0.5472 & 0.5738 \\
\cline{2-4}
& MuRIL & 0.5802 & 0.5463\\
\hline
\multirow{6}{4em}{Dense} & mbert-uncased & 0.5916 & 0.4473 \\
\cline{2-4}
& mbert-cased & 0.5946 & 0.4527 \\
\cline{2-4}
& indic-bert & 0.5609 & 0.5785 \\
\cline{2-4}
& xlm-roberta-base & 0.5481 & 0.3936 \\
\cline{2-4}
& distilmbert-cased & 0.6034 & 0.5926 \\
\cline{2-4}
& MuRIL & 0.5504 & 0.5291\\

\hline
\end{tabular}
\caption{Weighted F1-scores of hope speech detection classifier models on Tamil dataset}
\label{tab8}
\end{center}

\end{table*} 
\subsubsection{CharacterBERT}
CharacterBERT (CharBERT) \cite{el-boukkouri-etal-2020-characterbert} is a variant of BERT \cite{devlin-etal-2019-bert} which uses CharacterCNN \cite{NIPS2015_250cf8b5} like ELMo \cite{peters-etal-2018-deep}, instead of relying on WordPieces \cite{Wu2016GooglesNM}. CharacterBERT is highly desired as it produces a single embedding for any input token which is more suitable than having an inconstant number of WordPiece vectors for each token. It furthermore replaces BERT from domain-specific wordpiece vocabulary and enables it to be more robust to noisy inputs.

We use the pretrained model \textbf{\textit{general-character-bert}\footnote{\url{https://github.com/helboukkouri/character-bert}}} which was pretrained on the same corpus of that of BERT, but with a different tokenization approach. A CharacterCNN module is used that produces word-level contextual representations and it can be re-adapted to any domain without needing to worry about the suitability of any wordpieces (Figure \ref{fig1}). This approach helps for superior robustness by approaching the character of the inputs.

\begin{table*}[]
\begin{tabular}{|l|r|r|r|r|r|}
\hline
Language & Hope-Speech & Not-hope speech & Other Language & Macro Avg & Weighted Avg \\ \hline
\multicolumn{6}{|c|}{\textbf{Precision}}                                                        \\ \hline
English    &   0.9464          &      0.6346           &     0.0000           &   0.5270        &    0.9193          \\
Malayalam  &   0.6540          &    0.9032             &     0.8941           &    0.8171       &   0.572           \\
Tamil      &     0.4824        &      0.5819           &      0.5709          &     0.5451      &     0.5403         \\ \hline
\multicolumn{6}{|c|}{\textbf{Recall}}                                                           \\ \hline
English    &      0.9781       &       0.4108          &          0.0000      &   0.4630       &     0.9293         \\
Malayalam  &     0.7113        &       0.9021          &     0.7525           &   0.7886        &  0.8534            \\
Tamil      &     0.2687        &         0.7812        &       0.6525         &     0.5675      &     0.5579         \\ \hline
\multicolumn{6}{|c|}{\textbf{F1-Score}}                                                         \\ \hline
English    &     0.9620        &      0.4987           &     0.0000           &    0.4869       &  0.9220            \\
Malayalam  &    0.6815         &      0.9026           &      0.8172          &      0.8004     &        0.8545      \\
Tamil      &      0.3452       &        0.6670         &        0.6090        &      0.5404     &        0.5207      \\ \hline
\end{tabular}
\caption{Classification report for our system models based on the results of test set}
\label{table:classification report}
\end{table*}

\subsubsection{ULMFiT}
Universal Language Model Fine-tuning, or ULMFiT, was a transfer learning method introduced to perform various NLP tasks \cite{howard-ruder-2018-universal}. Training of ULMFiT involves pretraining the general language model on a Wikipedia-based corpus, fine-tuning the language model on a target text, and finally, fine-tuning the classifier on the target task. Discriminative fine-tuning is applied to fine-tune the model as different layers capture the different extent of information. It is then trained using the learning rate scheduling strategy, Slanted triangular learning rates (STLR), where the learning rate increases initially and then drops. Gradual unfreezing is used to fine-tune the target classifier rather than training all layers at once, which might lead to catastrophic forgetting.

Pretrained model, \textbf{AWD-LSTM} \cite{merity2017regularizing} with 3 layers and 1150 hidden activation per layer and an embedding size of 400 is used as the language model for the English dataset. Adam optimizer with $\beta_1=0.9$ and $\beta_2=0.99$ is implemented. Later, the start and end learning rates are set to \emph{1e-8 and 1e-2} respectively and fine-tuned by gradually unfreezing the layers to produce better results. Dropouts with a multiplier of 0.5 were applied. 

\subsubsection{XLM-RoBERTa}
XLM-RoBERTa \cite{ruder-etal-2019-unsupervised} is a pre-trained multilingual language model to execute diverse NLP transfer tasks. It's trained on over 2TB of filtered CommonCrawl data in 100 different languages. It was an update to the XLM-100 model \cite{lample2019crosslingual} but with increased training data. As it shares the same training routine with the RoBERTa model, "RoBERTa" was included in the name. \textbf{\textit{xlm-roberta-base}} with 12 layers, 768 hidden, 12 heads, and 270M parameters were used. It is fine-tuned for classifying code-mixed Tamil and Malayalam datasets.

\subsubsection{MuRIL}
MuRiL\footnote{\url{https://tfhub.dev/google/MuRIL/1}} was introduced by Google Research India to enhance Indian NLU (Natural Language Understanding). The model has a BERT based architecture trained on 17 Indian languages with Wikipedia, Common Crawl\footnote{\url{http://commoncrawl.org/the-data/}}, PMINDIA\footnote{\url{http://lotus.kuee.kyoto-u.ac.jp/WAT/indic-multilingual/index.html}} and Dakshina\footnote{\url{https://github.com/google-research-datasets/dakshina}} datasets. MuRIL is trained on translation and transliteration segment pairs which give an advantage as the transliterated text is very common in social media. It is used for the Malayalam and Tamil datasets.

\subsubsection{IndicBERT}

IndicBERT \cite{kakwani2020indicnlpsuite} is an ALBERT model pretrained on 12 major Indian languages with a corpus of over 9 billion tokens. It performs as well as other multilingual models with considerably fewer parameters for various NLP tasks. It's trained by choosing a single model for all languages to learn the relationship between languages and understand code-mixed data. \textbf{\textit{ai4bharat/indic-bert}} model was employed for the Tamil and Malayalam task.

\section{Results}

In this section, we have compared the F1-scores of our transformer-based models to successfully classify social media comments/posts into hope speech or not hope speech and detect the usage of other languages if any. We have tabulated the weighted average F1-scores of our various models for validation and test dataset for English, Malayalam and Tamil languages in tables \ref{tab6}, \ref{tab7} and \ref{tab8} respectively.

Table \ref{tab6} demonstrates that the character-aware model CharacterBERT performed exceptionally well for the validation dataset. It beat ULMFiT \cite{howard-ruder-2018-universal} by a mere difference of 0.0012, but other BERT-based models like BERT \cite{devlin-etal-2019-bert} with dense and BiLSTM architecture, ALBERT \cite{chiang-etal-2020-pretrained}, DistilBERT \cite{unknown5} and RoBERTa \cite{liu2020roberta} by about a percent. This promising result shown by character-bert for the validation dataset made it our best model. Unfortunately, few models managed to perform better than it for the test dataset. The considerable class imbalance of about 2,484 hope to 25,940 not hope comments and the interference of comments in other languages have significantly affected the results.

Similar transformer-based model trained on multilingual data was used to classify Malayalam and Tamil datasets. Models like multilingual BERT, XLM-RoBERTa \cite{ruder-etal-2019-unsupervised}, MuRIL, IndicBERT \footnote{https://indicnlp.ai4bharat.org/indic-bert/} and DistilBERT multilingual with both BiLSTM and Dense architectures. mBERT (Multilingual BERT) uncased with BiLSTM concatenated to it outperformed the other models for the Malayalam validation dataset and continued its dominance for the test data as well. 

The data distribution for the Tamil dataset seemed a bit balanced with an approximate ratio of 4:5 between hope and not-hope. mBERT cased with BiLSTM architecture appeared to be the best model with an F1-score of 0.6183 for validation but dropped drastically by 8\% for the test data. We witnessed a considerable fall in the scores of other models like mBERT and XLM-RoBERTa with linear layers of up to 15\%. 

Multilingual comments experience an enormous variety of text as people tend to write in code-mixed data and other non-native scripts which are inclined to be mispredicted. A variation in the concentration of such comments between train, validation and test can result in a fluctuation in the test results. The precision, recall and F1-scores of CharacterBERT, mBERT-uncased, and mBERT-cased are tabulated under English, Malayalam, and Tamil respectively, as shown in Table \ref{table:classification report}. They were the best performing models on the validation set.

\section{Conclusion}
During these unprecedented times, there is a need to detect positive, enjoyable content on social media in order to help people who are combating depression, anxiety, melancholy, etc.  This paper presents several methodologies that can detect hope in social media comments. We have traversed through transfer learning of several state-of-the-art transformer models for languages such as English, Tamil, and Malayalam. Due to its superior fine-tuning method, ULMFiT achieves an F1-score of 0.9356 on English data. We observe that mBERT achieves 0.8545 on Malayalam test set and distilmBERT achieves 0.5926 weighted F1-score on Tamil test set.

\bibliography{anthology,eacl2021}
\bibliographystyle{acl_natbib}

\end{document}